# PROBABILISTIC INFERENCE AND PROBABILISTIC REASONING*
Henry E. Kyburg, Jr. University of Rochester

## 1. Probabilistic Reasoning

Uncertainty enters into human reasoning and inference in at least two ways. It is reasonable to suppose that there will be roles for these distinct uses of uncertainty also in automated reasoning.

One role for uncertainty concerns choices among alternative actions. For good reasons, a conception of uncertainty that did *not* satisfy the probability axioms, but that was used for computing expected utilities, would lead the agent into decisions under which the agent would come out on the short end in every eventuality (Ramsey, 1950). Uncertainty, construed in this way, is what we need for computing the expectations that are fed into decision rules, the most common and persuasive of which is the rule to maximize expected utility.

Uncertainty need not be represented by a single classical probability function; if uncertainties are represented in some more general fashion, what the Dutch Book argument shows is that there should exist some classical probability function that is compatible with that more general representation. It has been argued (Levi, 1980; Kyburg, 1987) that the most general form of representation for these uncertainties is that of a set of classical probability functions, having a convex hull, defined over an algebra of propositions. Such a representation includes as special cases belief functions (Shafer, 1976) and most interval representations of uncertainty. Less general forms of convexity, for example, in which a set of parameterized probability functions is taken to be convex over values of the parameter, are also of interest. Manipulating probability and utility representations, seeing which ones are consistent with which, or which imply which, constitutes one form of probabilistic reasoning. It is this kind of reasoning that Nilsson (1986) considers to be the appropriate subject matter of probabilistic logic.

In addition to merely representing uncertainty and employing it in decision theory, we are concerned with how uncertainties are modified or updated in response to evidence. The classical way of doing this, for classical probabilities, is by means of Bayes' theorem: if $E$ becomes known, is accepted as evidence, then the new or updated probability $P'$ of any statement $H$ in our algebra becomes the old probability of $H$ multiplied by the ratio of the likelihood of $H$ on $E$ to the old probability of $E$:

$$P'(H) = P(H) * (P(E|H) / P(E))$$

221

This is called confirmational conditionalization. A more general procedure is Jeffrey Conditionalization (Jeffrey, 1965), which applies when we undergo some experience or make some observation whose import is exactly to lead us to shift some probability from $P(E)$ to $P'(E)$. Then the new probability of every other statement becomes:

$$P'(H) = P(H|E) * P'(E) + P(H|\tilde{E})(1 - P'(E))$$

Confirmational conditionalization can be extended to the more general approach that represents uncertainty by convex sets of classical probabilities: it can be shown that if each classical probability function in a convex set of probability functions is updated by conditionalizing on the evidence $E$, the result will be a new convex set of classical probability functions, provided $E$ does not have zero probability on all the original probability functions (Kyburg, 1987).

There are other ways in which one might want to update probabilities than by conditionalization. Certain forms of direct inference, in which probabilities are derived from knowledge of statistics are one possibility. But while any of these procedures has a perfect right to be called 'probabilistic reasoning,' they are not what I mean by probabilistic inference.

## 2. Probabilistic Reasoning

In inference, in general, one begins with certain statements or propositions (representations of states of affairs), premises, and goes through a process that leads to another statement, the conclusion. From "Tosses of this coin are independent and heads occur half the time," we infer, not probabilistically, but deductively, that if we know that heads has occurred on the first toss of a pair, then the probability that heads also occurred on the second toss of the pair is a half. We infer, not probabilistically, but deductively, that triples of tosses consisting of three heads occur an eighth of the time. We infer, not probabilistically, but deductively, that if our sample of $n$ from $P$ is random with respect to $R$, then the probability is at least 0.91 that the proportion of $P$ that are $R$ lies within $3/(4n)^{1/2}$ of the observed sample population.

In ordinary deductive logic, the process of inference is such as to preserve truth: if the premises are true, so is the conclusion. Note that the probabilistic reasoning mentioned above fits into this deductive pattern.

A sound deductive argument is one in which the premises are true and the argument valid. In most applications of deductive inference, the premises may be warranted, firmly believed, accepted as practically certain. Sometimes the premises are merely accepted hypothetically, or for the purposes of argument. To have reason to believe a valid deductive argument is sound is precisely to have reason to believe that its premises are true.

Often a valid deductive inference provides warrant for accepting its conclusion. But not always: confronted with the validity of the inference: "All swans are white; Sam is a black Australian swan; Therefore Sam is white," we opt for the rejection of the universal premise rather than for the (inconsistent) inclusion of the consequence.

Spelling out the conditions under which valid deductive inference provides warrant for its conclusions is not trivial. It is not trivial because it depends on spelling out the justification for the premises of a deductive argument.

### 3. Ampliative Inference.

What is controversial is whether or not there is any form of inference other than deductive inference. There is a tradition in philosophy that considers "inductive" or "ampliative" inference. There is a new tradition, in artificial intelligence, according to which inference proceeds by rules that may not be truth-preserving in the sense that an addition to our knowledge base may require the rejection of a once acceptable conclusion. Such non-monotonic rules will lead to truth for the most part.

David Israel (1986) suggests that there is no other logic, but that "real" inference proceeds in non-deductive ways. This is just what inductive or ampliative inference, or more recently scientific inference, or non-monotonic infernce, has been concerned with. Whether or not it is to be called a "logic" seems unimportant.

In artificial intelligence, this search is to be found in the search for representations of non-monotonic reasoning, such as circumscription, non-monotonic logics, default logics, and the like. We want to be able to infer that Tweety can fly. Since the kinds of inference under investigation do not preserve truth, we have to be able to back up: if we enlarge the premiss set, we may have to shrink the conclusion set. Non-monotonic inference is not generally taken to be probabilistic, but work on non-monotonic logic suggests that there is interest in inference rules -- that is, rules that lead from premises to the acceptance of a conclusion -- that need not be truth preserving. Many people want to be able to detach conclusions from their premises. Not all approaches to non-monotonic logic allow full detachment; de Kleer's ATMS (de Kleer, 1986), for example, requires that tags reflecting the assumptions used in an inference be

223

carried along with the conclusions.

## 4. Why Accept?

There may be a sense in which, when I choose to bet at even money on heads rather than on tails, I may be said to be "accepting" the hypothesis that heads will come up. This is not an interesting sense of acceptance. Decision theory can account for such usages. I shall speak of accepting $S$ only when I am adding $S$ to my knowledge base, using $S$ as a premise in deductive arguments I take to be sound, using $S$ as evidence in inductive arguments, etc.

Despite the fact that some people are interested in non-deductive inference, we may still sensibly ask why they should be: Why should we accept any statements that are not (say) mathematical or logical truths? It might be thought that we couldn't use conditionalization for updating without acceptance: after all, when we update on evidence $E$, we take the probability of $E$ to be one. And once a statement has a probability of 1 (or of 0) that probability can never be changed by conditionalization. But there are other ways to handle updating: Jeffrey's rule, for example, or various net-propagation procedures, such as Pearl's (Pearl, 1986).

In principle, there is no reason that human or machine knowledge in a certain domain should not be represented by a complete algebra of statements and a classical probability measure (or a set of classical probability measures) over them, in which no empirical statement ever receives a probability of 0 or 1. Such a system would have no need for probabilistic rules of inference; it would employ only deductive rules for manipulating probabilities.

While such a system would be conceptually simple, it would not reflect the way in which people function epistemologically. We are willing to assert categorically a large number of propositions that could conceivably be false. Perhaps more to the point, the progress of science depends on the categorical acceptance of such statements as: measurement $m$ is not in error by more than $\Delta$. The efforts of engineers in designing a tool or a product depend on their ability to take as given such possibly erroneous statements as those concerning strength of materials, conductivity, ... Such statements we take as premises in deductive arguments, whose conclusions we therefore also accept. We are willing to take such statements as evidence -- e.g., I take as evidence the statement that about 50% of tosses result in heads; it is relative to this evidence that I assign a probability of heads on the next toss of about .5.

Our empirical scientific knowledge is expressed, not in probabilities (for the most part) but in categorical statements. Is this an idealization?

224

Perhaps. But the fact that it is such a pervasive idealization suggests that there is a reason for it, and that we might be well advised to emulate it in our formal representations.

## 5. Models of Probabilistic Inference.

In testing a statistical hypothesis, the standard goal is to devise a rule that will <u>erroneously reject</u> that hypothesis no more than $\alpha$ of the time. Such a test will lead you to a false rejection no more frequently than $\alpha$ (Lehman, 1959). Of course $\alpha$ is a free parameter; but we choose $\alpha$ to be small enough that the possibility of making this sort of error in a given context does not worry us. The size $\alpha$ we choose reflects how seriously we take the mistake in question. If it is very serious, we want to be very sure (but we <u>can't</u> ask for a guarantee) that it won't happen.

It is very bad form to say of a hypothesis that has been rejected at the level $\alpha$ that the probability is at most $\alpha$ that it was falsely rejected. But as Birnbaum has pointed out (Birnbaum, 1969), while we can learn not to say such things, it is hard to know what else to think.

Consider the simplest and most elegant of all forms of statistical inference: you have a normally distributed quantity $X$, but you don't know the parameters of its distribution. Nevertheless, since you know that it is normally distributed, you <u>know</u> the distribution of the quantity $t = N^{1/2}(x - \mu)(s^{-1})$, where $x$ and $s$ are the sample mean and standard deviation, and $\mu$ is the unknown population mean. Knowing the distribution of $t$, you can therefore compute the probability, for example, that

$$x - ts/N^{1/2} < \mu < x + ts/N^{1/2}.$$

If you pick some probability level that makes you feel comfortable under the circumstances, and you are indifferent between over and underestimating $\mu$, then you will have an exact interval estimate of the unknown mean $\mu$, indexed by $f_\rho$: a level of fiducial probability or practical certainty.

Note that this inference is non-monotonic: on observing a further sample from the population it may well be that some different interval for $\mu$ will be acceptable at the index $f_\rho$.

Or consider the most common form of confidence interval inference: you have a binomial population with an unknown parameter $r$; you draw a sample from the population, and <u>observe</u> a relative frequency $f$; you construct a class of intervals $(\rho_l, \rho_u)$ such that <u>whatever</u> the true

225

value of $r$ may be, the probability is at least $p$ that the sample frequency will fall in the corresponding interval. But it will have done this *if and only if* $r$ lies between a certain maximum and a certain minimum value. These values determine what is called a <u>confidence interval</u>, and in particular, a $100p$% confidence interval, since its limits require the specification of an acceptable $p$.

In philosophy, Levi (1967) is concerned with the circumstances under which one ought to add a hypothesis to one's corpus of knowledge. The famous *Rule A* for doing so involves, in addition to the probability of the hypothesis, a measure of the epistemic content of the hypothesis, and a further parameter $q$, which varies from 0 to 1 and is called an index of caution.

In artificial intelligence Matthew Ginsberg (1985) applies a technique much like that of binomial confidence interval inference (the main difference being that he uses a rougher approximation) to the problem of inferring an interval characterizing the reliability of a default rule in non-monotonic logic. In order to do this, he finds it necessary to introduce a parameter $g$, analogous to the fiducial parameter $f_p$, which he calls "gullibility."

### 6. Probabilistic Acceptance.

The simplest and most natural idea for acceptance in AI is just to accept those statements whose probability exceeds a certain critical number. This number may have to be changed to reflect different circumstances -- it will be context dependent -- but so, we may suppose, are $\alpha$, $g$, $q$, $p$, and $f_p$ context dependent.

In what way is acceptance level context dependent? One natural answer is that acceptance level depends on what is, or might be expected to be, at stake. If the range of stakes we are contemplating is limited -- for example, it can't be more than 10 to 1 -- then probabilities greater than .9 are behaviorally indistinguishable from probabilities of 1, and probabilities of less than .1 are indistinguishable from probabilities of 0.

It also follows from these considerations that probabilities larger than the level of acceptance, or smaller than 1 - the level of acceptance, are just not significant as probabilities. It makes no sense to bet at odds of 1000:1 on a statement that gets its probability from a statistical statement whose acceptance level is only .99. If you're only 99% sure that the coin lands heads between .48 and .52 of the time, you should not be willing to bet at odds of a thousand to one than in 12 tosses you won't get 12 heads.

Most of the acceptance rules mentioned above run afoul of the

226

lottery paradox (Kyburg, 1961). That is, each of a set of statements $S_i$ (e.g., "ticket $i$ will not win the lottery") may be probable enough to be accepted, and at the same time they may jointly contradict other accepted statements (e.g., "there will be a winner"). An exception is the acceptance principle advocated by Levi, which links acceptance to expected epistemic utility; only statements demonstrably consistent with what you have already accepted are candidates for future acceptance.

How serious the lottery paradox is depends on what other machinery you have. It is not deadly if you limit yourself to a probabilistic rule of acceptance. It will follow that any logical consequence of a single statement in your corpus of knowledge should also be in it; but it will not follow that every consequence of the set of sentences in your corpus will also be in it. The latter would indeed lead to a hopeless sort of inconsistency; the former would not. If the size of the lottery is adjusted to my level of acceptance, I will answer your question about whether ticket $i$ will win with a categorical "no." But I will answer your question about whether it is true that neither ticket $i$ nor ticket $j$ will win by saying, "I don't know; but the probability is thus and so."

This seems not unreasonable. Look at the matter another way: given a deductive argument from premises $P_1, ..., P_n$ to a conclusion $C$, consider whether the argument obligates you to accept $C$. It seems natural to say that more is required than merely that each of the premises be accepted; I must also be willing to accept the conjunction of the premises.

This feature might even be advantageous in A I. There is surely an epistemic difference between a conclusion reached in one step from a single premise, and a conclusion that requires a large number of steps and premises. This difference disappears if the acceptability of the single premise of the first agument is no greater than that of the conjunction of all the premises in the second argument. A purely probabilistic rule of acceptance automatically reflects this fact.

## 7. Conclusion.

It is important to distinguish probabilistic reasoning from probabilistic inference. Probabilistic reasoning may concern the manipulation of probabilities in the context of decision theory, or it may involve the updating of probabilities in the light of new evidence via Bayes' theorem or some other procedure. Both of these operations are essentially deductive in character.

Contrasted with these procedures of manipulating or computing with probabilities, is the use of probabilistic rules of inference: rules that lead from one sentence (or a set of sentences) to another sentence, but do so in a way that need not be truth preserving.

Instances of such rules are those represented by circumscription, non-monotonic logic, default rules, etc. as well as several classes of inference rules associated with statistics, and some rules discussed by philosophers.

The simplest probabilistic rule of inference -- a high probability rule -- has some curious consequences (the lottery paradox), but it does not seem that these consequences need interfere with the useful application of the rule. Other rules also lead to the lottery paradox.

* Research underlying this paper has been partially supported by the Signals Warfare Center of the U. S. Army.